\documentclass{article}
\usepackage{ijcai21}

\usepackage{times}
\usepackage{soul}
\usepackage{url}
\usepackage[hidelinks]{hyperref}
\usepackage[utf8]{inputenc}
\usepackage[small]{caption}
\usepackage{amsmath,amssymb,amsfonts}
\usepackage{booktabs}
\usepackage{enumitem}
\usepackage{multirow}
\usepackage{subcaption}
\usepackage{graphicx}
\usepackage{xcolor}
\newcommand{\retain}{{RETAIN}}
\newcommand{\deepr}{{Deepr}}
\newcommand{\gram}{{GRAM}}

\newcommand{\dipole}{{Dipole}}
\newcommand{\timeline}{{Timeline}}
\newcommand{\notes}{{LR$_{\text{notes}}$}}
\def\modelname{{CGL}}
\newcommand{\medgcn}{{MedGCN}}

\title{Collaborative Graph Learning with Auxiliary Text for Temporal Event Prediction in Healthcare} % \\

\author{
Chang Lu$^1$\and
Chandan K. Reddy$^2$\and
Prithwish Chakraborty$^3$\and
Samantha Kleinberg$^1$\and
Yue Ning$^1$
\affiliations
$^1$Department of Computer Science, Stevens Institute of Technology\\
$^2$Department of Computer Science, Virginia Tech\\
$^3$IBM Research
\emails
\{clu13, samantha.kleinberg, yue.ning\}@stevens.edu,
reddy@cs.vt.edu,
prithwish.chakraborty@ibm.com
}

\begin{document}

\maketitle

\begin{abstract}
Accurate and explainable health event predictions are becoming crucial for healthcare providers to develop care plans for patients. The availability of electronic health records (EHR) has enabled machine learning advances in providing these predictions. However, many deep learning based methods are not satisfactory in solving several key challenges: 1) effectively utilizing disease domain knowledge; 2) collaboratively learning representations of patients and diseases; and 3) incorporating unstructured text. To address these issues, we propose a collaborative graph learning model to explore patient-disease interactions and medical domain knowledge. Our solution is able to capture structural features of both patients and diseases. The proposed model also utilizes unstructured text data by employing an attention regulation strategy and then integrates attentive text features into a sequential learning process. We conduct extensive experiments on two important healthcare problems to show the competitive prediction performance of the proposed method compared with various state-of-the-art models. We also confirm the effectiveness of learned representations and model interpretability by a set of ablation and case studies.
\end{abstract}

\section{Introduction}
Electronic health records (EHR) consist of patients' temporal visit information in health facilities, such as medical history and doctors' diagnoses. The usage and analysis of EHR not only improves the quality and efficiency of in-hospital patient care but also provides valuable data sources for researchers to predict health events, including diagnoses, medications, and mortality rates, etc. A key research problem is improving prediction performance by learning better representations of patients and diseases so that improved risk control and treatments can be provided. There have been many works on this problem using deep learning models, such as recurrent neural networks (RNN)~\cite{choi2016doctor}, convolutional neural networks (CNN)~\cite{Phuoc2017deepr}, and attention-based mechanisms~\cite{ma2017dipole}. However, several challenges remain in utilizing EHR data and interpreting models:

\begin{figure}
  \centering
  \includegraphics[width=0.82\linewidth]{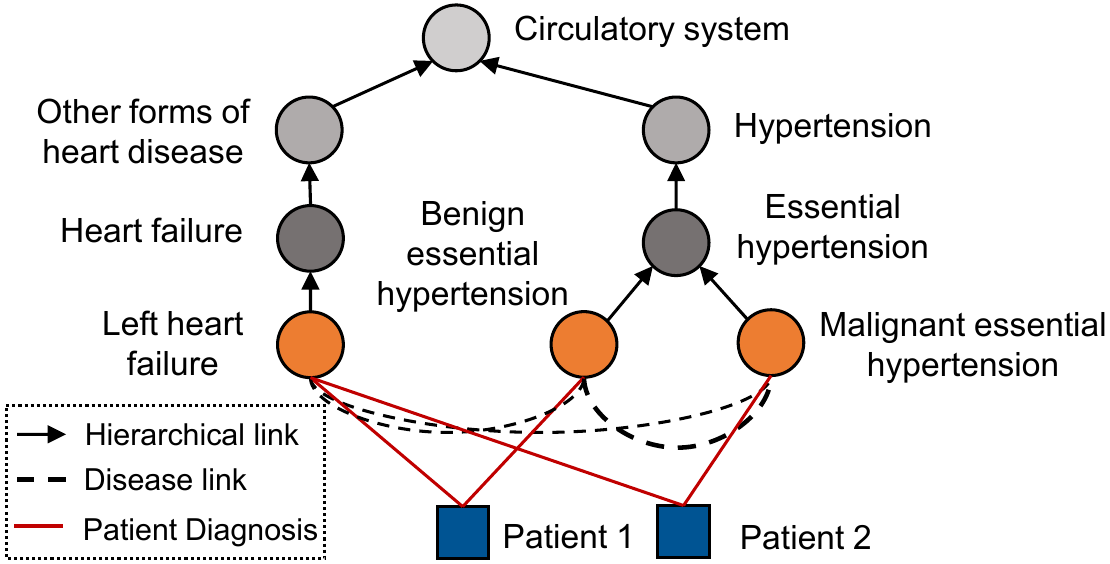}
  \caption{An example of the hierarchical structure of the ICD-9-CM system, disease interaction, and patient diagnosis.}
  \label{fig:hierarchical_example}
\end{figure}
\begin{enumerate}[leftmargin=*]
    \item \textbf{Effectively utilizing the domain knowledge of diseases}. Recently, graph structures are being adopted~\cite{choi2017gram} using disease hierarchies, where diseases are classified into various types at different levels. For example, \figurename{~\ref{fig:hierarchical_example}} shows a classification of two forms of hypertension and one form of heart failure. One problem is that existing works~\cite{choi2017gram,shang2019pre} only consider the vertical relationship between a disease and its ancestors (hierarchical link). However, they ignore horizontal disease links that can reflect disease complications and help to predict future diagnoses.
    \item \textbf{Collaboratively learning patient-disease interactions.} Patients with the same diagnoses may have other similar diseases (patient diagnosis in \figurename~{\ref{fig:hierarchical_example}}). Existing approaches ~\cite{choi2017gram,ma2017dipole} treat patients as independent samples by using diagnoses to represent patients, but they fail to capture patient similarities, which help in predicting new-onset diseases from other patients.
    \item \textbf{Incorporating unstructured text.} Unstructured data in EHR including clinical notes contain indicative features such as physical conditions and medical history. For~example, a note: ``\textit{The patient was intubated for respiratory distress and increased work of breathing. He was also hypertensive with systolic in the 70s}'' indicates that this patient has a history of respiratory problems and hypertension. Most models~\cite{choi2016retain,bai2018interpretable} do not fully utilize such data. This often leads to unsatisfactory prediction performance and lack of interpretability.
\end{enumerate}

To address these problems, we first present a hierarchical embedding method for diseases to utilize medical domain knowledge. Then, we design a collaborative graph neural network to learn hidden representations from two graphs: a {\it patient-disease observation graph} and a {\it disease ontology graph}. In the observation graph, if a patient is diagnosed with a disease, we create an edge between this patient and the disease. The ontology graph uses weighted ontology edges to describe horizontal disease interactions. Moreover, to learn the contributions of keywords for predictions, we design a TF-IDF-rectified attention mechanism for clinical notes which takes visit temporal features as context information. Finally, combining disease and text features, the proposed model is evaluated on two tasks: predicting patients' future diagnoses and heart failure events. The main contributions of this work are summarized as follows:
\begin{itemize}
    \item We propose to collaboratively learn the representations of patients and diseases on the observation and ontology graphs. We also utilize the hierarchical structure of medical domain knowledge and introduce an ontology weight to capture hidden disease correlations. 
    \item We integrate structured information of patients' previous diagnoses and unstructured information of clinical notes with a TF-IDF-rectified attention method. 
    It allows us to regulate attention scores without any manual intervention and alleviates the issue of using attention as a tool to audit a model~\cite{jain-wallace-2019-attention}.
    \item We conduct extensive experiments and illustrate that the proposed model outperforms state-of-the-art models for prediction tasks on MIMIC-III dataset. We also provide detailed analysis for model predictions.
\end{itemize}
\section{Related Work}
\label{sec:related_works}
Deep learning models, especially RNN models, have been applied to predict health events and learn representations of medical concepts.  DoctorAI~\cite{choi2016doctor} uses RNN to predict diagnoses in patients' next visits and the time duration between patients' current and next visits. RETAIN~\cite{choi2016retain} improves the prediction accuracy through a sophisticated attention process on RNN. Dipole~\cite{ma2017dipole} uses a bi-directional RNN and attention to predict diagnoses of patients' next visits. Both Timeline~\cite{bai2018interpretable} and  ConCare~\cite{concare2020} utilize time-aware attention mechanisms in RNN for health event predictions. However, RNN-based models regard patients as independent samples and ignore relationships between diseases and patients which help to predict diagnoses for similar patients.

Recently, graph structures are adopted to explore medical domain knowledge and relations of medical concepts.  GRAM~\cite{choi2017gram} constructs a disease graph from medical knowledge. MiME~\cite{choi2018mime} utilizes connections between diagnoses and treatments in each visit to construct a graph. GBERT~\cite{shang2019pre} jointly learns two graph structures of diseases and medications to recommend medications. It uses a bi-directional transformer to learn visit embeddings. MedGCN~\cite{mao2019medgcn} combines patients, visits, lab results, and medicines to construct a heterogeneous graph for medication recommendations. GCT~\cite{gct_aaai20} also builds graph structures of diagnoses, treatments, and lab results. However, these models only consider disease hierarchical structures, while neglecting disease horizontal links that reflect hidden disease complications. As a result, prediction performance is limited.

In addition, CNN and Autoencoder are also adopted to predict health events. DeepPatient~\cite{miotto2016deep} uses an MLP as an autoencoder to rebuild features in EHR. Deepr~\cite{Phuoc2017deepr} treats diagnoses in a visit as words to predict future risks such as readmissions in three months. AdaCare~\cite{ma2020adacare} uses multi-scale dilated convolution to capture dynamic variations of biomarkers over time. However, these models neither consider medical domain knowledge nor explore patient similarities as discussed.

In this paper, we explore disease horizontal connections using a disease ontology graph. We collaboratively learn representations of both patients and diseases in their associated networks. We also design an attention regulation strategy on unstructured text features to provide quantified contributions of clinical notes and interpretations of prediction results.

\section{Methodology}\label{sec:methodology}

\subsection{Problem Formulation}
An EHR dataset is a collection of patient visit records. Let $\mathcal{C} = \{c_1, c_2, \dots, c_{|\mathcal{C}|}\}$ be the entire set of diseases represented by medical codes in an EHR dataset, where $|\mathcal{C}|$ is the medical code number. Let $\mathcal{N} = \{\omega_1, \omega_2, \dots, \omega_{|\mathcal{N}|}\}$ be the dictionary of clinical notes, where $|\mathcal{N}|$ is the word number.

\paragraph{EHR dataset.} An EHR dataset is given by $\mathcal{D} = \{ \gamma_u | u \in \mathcal{U} \}$ where $\mathcal{U}$ is the collection of patients in $\mathcal{D}$ and $\gamma_u = (V^u_1,V^u_2, \dots, V^u_T)$ is a visit sequence of patient $u$. Each visit $V^u_t = \{C^u_t, N^u_t \}$ is recorded with a subset of medical codes $C^u_t \subset \mathcal{C}$, and a paragraph of clinical notes $ N^u_t \subset \mathcal{N}$ containing a sequence of $|N^u_t|$ words.

\paragraph{Diagnosis prediction.} Given a patient $u$'s previous $T$ visits, this task predicts a binary vector $\mathbf{\hat{y}} \in \{0, 1\}^{|\mathcal{C}|}$ which represents the possible diagnoses in $(T + 1)$-th visit. $\mathbf{\hat{y}}_i =1$ denotes $c_i $ is predicted in $ C^u_{T + 1}$.

\paragraph{Heart failure prediction.} Given a patient $u$'s previous $T$ visits, this task predicts a binary value $\hat{y} \in \{0, 1\}$. $\hat{y} = 1$~denotes that $u$ is predicted with heart failure\footnote{The medical codes of heart failure start with 428 in ICD-9-CM.}  in $(T + 1)$-th visit.

In the rest of this paper, we drop the superscript $u$ in $V^u_t, C^u_t$, and $N^u_t$ for convenience unless otherwise stated. 

\subsection{The Proposed Model}
\begin{figure}
  \centering
  \includegraphics[width=\linewidth]{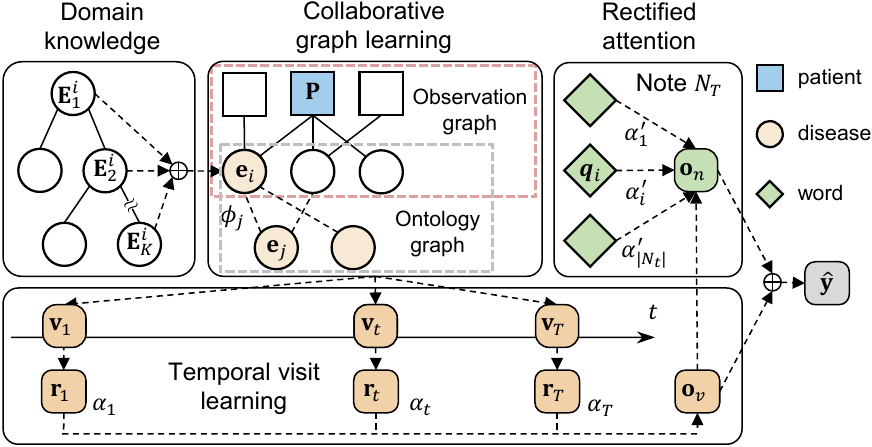}
  \caption{An overview of the proposed model. The graph learning first learns disease hidden features with two collaborative graphs: an observation graph and an ontology graph, based on the hierarchical embedding from medical domain knowledge. Then an RNN is designed to learn temporal information of visit sequences. Rectified Attention mechanism encodes clinical notes with the guide of TF-IDF and uses the visit representation as an attention context vector to integrate structured visit records and unstructured clinical notes.}
  \label{fig:proposed_model}
\end{figure}

In this section, we propose a \textbf{C}ollaborative \textbf{G}raph \textbf{L}earning model, \textbf{\modelname}. An overview of {\modelname} is shown in \figurename{~\ref{fig:proposed_model}}.

\subsubsection{Hierarchical Embedding for Medical Codes}
ICD-9-CM is an official system of assigning codes to diseases. It hierarchically classifies medical codes into different types of diseases in $K$ levels. This forms a tree structure where each node has only one parent. Note that most medical codes in patients' visits from EHR data are leaf nodes. However, a patient can be diagnosed with a higher level disease, i.e., non-leaf node. Therefore, we recursively create virtual child nodes for each non-leaf node to pad them into virtual leaf nodes. We assume there are $n_k$ nodes at each level $k$ (smaller $k$ means higher level in the hierarchical structure).

We create an embedding tensor $\{ \mathbf{E}_k \}_{k \in [1, 2, \dots, K]}$ for nodes in the tree. $\mathbf{E}_k \in \mathbb{R}^{n_k \times d_c}$ is the embedding matrix for nodes in level $k$, and $d_c$ is the embedding size. For a medical code $c_i$ as a leaf node, we first identify its ancestors in each level $k=[1, 2, \dots, K-1]$ in the tree and select corresponding embedding vectors from $\{\mathbf{E}_k\}$. Then, the hierarchical embedding $\boldsymbol{\mathbf{e}}_i \in \mathbb{R}^{Kd_c}$ of $c_i$ is calculated by concatenating the embeddings in each level:  $\boldsymbol{\mathbf{e}}_i = \mathbf{E}_1^i \oplus \mathbf{E}_2^i~\oplus~, \dots, \oplus~\mathbf{E}_K^i$, where $\oplus$ denotes the concatenation. We use $\mathbf{E} \in \mathbb{R}^{|\mathcal{C}| \times Kd_c}$ to represent medical codes after hierarchical embedding.

\subsubsection{Graph Representation}
In visit records, specific diagnosis co-occurrences could reveal hidden similarities of patients and diseases. We explore such relationship by making the following hypotheses:
\begin{enumerate}[leftmargin=*]
    \item \textbf{Diagnostic similarity of patients}. If two patients get diagnosed with the same diseases, they tend to have diagnostic similarities and get similar diagnoses in the future.
    \item \textbf{Medical similarity of diseases}.
    If two diseases belong to the same higher-level disease, they might have medical similarities such as symptoms, causes, and complications.
\end{enumerate}

Based on these hypotheses, we construct a collaborative graph $\mathcal{G} = \{ \mathcal{G}_{\mathcal{U}\mathcal{C}}, \mathcal{G}_{\mathcal{C}\mathcal{C}} \}$ for patients and medical codes. $\mathcal{G}_{\mathcal{U}\mathcal{C}}$ is the patient-disease \textbf{\textit{observation graph}} built from EHR data. Its nodes are patients and medical codes. We use a patient-code adjacency matrix $\mathbf{A}_{\mathcal{U}\mathcal{C}} \in \{0, 1\}^{|\mathcal{U}|\times|\mathcal{C}|}$ to represent $\mathcal{G}_{\mathcal{U}\mathcal{C}}$. Given patient $u$, if $u$ is diagnosed with a code $c_i$ in a previous visit, we add an edge $(u, c_i)$ and set $\mathbf{A}_{\mathcal{U}\mathcal{C}}[u][i] = 1$. $\mathcal{G}_{\mathcal{C}\mathcal{C}}$ is the \textbf{\textit{ontology graph}}. Its nodes are medical codes. To model horizontal links of two medical codes (leaf nodes), we create a code-code adjacency matrix $\mathbf{A}'_{\mathcal{C}\mathcal{C}} \in \mathbb{N}^{|\mathcal{C}| \times |\mathcal{C}|}$. If two medical codes $c_i$ and $c_j$ have their lowest common ancestor in level $k$, we add an ontology edge $(c_i, c_j)_k$ and set $\mathbf{A}'_{\mathcal{C}\mathcal{C}}[i][j] = k$. This process is based on the idea that two medical codes with a common ancestor in lower levels of the hierarchial graph of ICD-9-CM should be similar diseases. Finally, we set $\mathbf{A}'_{\mathcal{C}\mathcal{C}}[i][i] = 0$ for all diagonal elements. Although $\mathbf{A}'_{\mathcal{C}\mathcal{C}}$ can reflect the hierarchical structure of medical codes, it is a dense matrix and generates a nearly complete ontology graph, which will cause a high complexity for graph learning. We further propose a disease co-occurrence indicator matrix $\mathbf{B}_{\mathcal{C}\mathcal{C}}$ initialized with all zeros. If two medical codes $c_i$ and $c_j$ appear in a patient's visit record, we set $\mathbf{B}_{\mathcal{C}\mathcal{C}}[i][j]$ and $ \mathbf{B}_{\mathcal{C}\mathcal{C}}[j][i]$ as 1. Then, we let $\mathbf{A}_{\mathcal{C}\mathcal{C}} = \mathbf{A}'_{\mathcal{C}\mathcal{C}} \odot \mathbf{B}_{\mathcal{C}\mathcal{C}}$ be a new adjacency matrix for $\mathcal{G}_{\mathcal{C}\mathcal{C}}$ to neglect disease pairs in $\mathbf{A}'_{\mathcal{C}\mathcal{C}}$ which never co-occur in EHR data. Here $\odot$ denotes element-wise multiplication. Finally, we not only create a sparse ontology graph for computational efficiency, but also focus on more common and reasonable disease connections in the ontology graph.

\subsubsection{Collaborative Graph Learning}
To learn hidden features of medical codes and patients, we design a collaborative graph learning method on the fact and ontology graphs. Instead of calculating patient embeddings with medical codes like DeepPatient~\cite{miotto2016deep}, we assign each patient an initial embedding. $\mathbf{P} \in \mathbb{R}^{|\mathcal{U}| \times d_p}$ is the embedding matrix of all patients with the size of $d_p$. Let $\mathbf{H}_p^{(0)} = \mathbf{P}, \mathbf{H}_c^{(0)} = \mathbf{E}$ and $\mathbf{H}_p^{(l)} \in \mathbb{R}^{|\mathcal{U}| \times d_p^{(l)}}, \mathbf{H}_c^{(l)} \in \mathbb{R}^{|\mathcal{C}| \times d_c^{(l)}}$ be the hidden features of patients and medical codes (i.e., inputs of $l$-th graph layer). We design a graph aggregation method to calculate the hidden features  of patients and medical codes in the next layer. First, we map the medical code features $\mathbf{H}_c^{(l)}$ into the patient dimension and aggregate adjacent medical codes from the observation graph ($\mathbf{A}_{\mathcal{U}\mathcal{C}}$) for each patient:
\begin{align}
    \label{eq:z_p}
    \mathbf{Z}_p^{(l)} = \mathbf{H}_p^{(l)} + \mathbf{A}_{\mathcal{U}\mathcal{C}}\mathbf{H}_c^{(l)}\mathbf{W_{\mathcal{C}\mathcal{U}}}^{(l)} \in \mathbb{R}^{|\mathcal{U}| \times d_p^{(l)}}.
\end{align}
Here $\mathbf{W_{\mathcal{C}\mathcal{U}}}^{(l)} \in \mathbb{R}^{d_c^{(l)} \times d_p^{(l)}}$ maps code embeddings to patient embeddings. For the ontology graph, if $c_i$, $c_j$ are connected in level $k$, we assign an ontology weight $\phi_j$ to $c_j$ when aggregating $c_j$ into $c_i$:
\begin{align}
    \phi_j\left(k\right) = \sigma\left({\mu}_j \times k + \theta_j\right).
\end{align}
Here $\sigma$ is the sigmoid function. $\mu_j, \theta_j \in \mathbb{R}$ are trainable variables for $c_j$. $\phi_j(k)$ is a monotonic function w.r.t. level $k$. This function enables the model to describe the horizontal influence of a disease on other diseases via assigning increasing or decreasing weights by levels. Let $\Phi \in \mathbb{R}^{|\mathcal{C}| \times|\mathcal{C}|}$ be the ontology weight matrix and $\mathbf{M}, \mathbf{\Theta} \in \mathbb{R}^{|\mathcal{C}|}$ be the collection of $\mu, \theta$. $\mathbf{H}_p^{(l)}$ is mapped into the medical code dimension and aggregated with adjacent patients from the observation graph:
\begin{align}
    \Phi = \sigma(\mathbf{M} \odot \mathbf{A}_{\mathcal{C}\mathcal{C}} + \mathbf{\Theta})\in \mathbb{R}^{|\mathcal{C}| \times|\mathcal{C}|}, \label{eq:phi}
\end{align}
\begin{align}
    \mathbf{Z}_c^{(l)} = \mathbf{H}_c^{(l)} + \mathbf{A}^{\top}_{\mathcal{U}\mathcal{C}}\mathbf{H}_p^{(l)}\mathbf{W_{\mathcal{U}\mathcal{C}}}^{(l)} + {\Phi\mathbf{H}_c^{(l)}} \in \mathbb{R}^{|\mathcal{C}| \times d_c^{(l)}}. \label{eq:z_c}
\end{align}
Here $\mathbf{W}_{\mathcal{U}\mathcal{C}} \in  \mathbb{R}^{d_p^{(l)} \times d_c^{(l)}}$ maps patient embeddings to code embeddings. Given that  $\mathbf{A}_{\mathcal{C}\mathcal{C}}$ stores the level where two diseases are connected, we use $\mathbf{A}_{\mathcal{C}\mathcal{C}}$ to compute $\Phi$. Finally, $\mathbf{H}_p^{(l)}$ and $\mathbf{H}_c^{(l)}$ of the next layer are calculated as follows:
\begin{align}
    \label{eq:layer_o}
    \mathbf{H}_{\{p, c\}}^{(l + 1)} = \text{ReLU}\left( \text{BatchNorm}\left( \mathbf{Z}_{\{p, c\}}^{(l)}\mathbf{W}^{(l)}_{\{p, c\}} \right)\right),
\end{align}
where $\mathbf{W}^{(l)}_{\{p, c\}}$ maps $\mathbf{Z}^{(l)}_{\{p, c\}}$ to the $(l + 1)$-th layer, and we use batch normalization to normalize features. In the $L$-th graph layers, we do not calculate $\mathbf{H}_{p}^{(L)}$ and only calculate $\mathbf{H}_{c}^{(L)}$ as the graph output, since the medical codes are required for further calculation. We let $\mathbf{H}_c = \mathbf{H}_{c}^{(L)} \in \mathbb{R}^{{|\mathcal{C}| \times d^{(L)}_c}}$ be the final embedding for medical codes.

\subsubsection{Temporal Learning for Visits}
Given a patient $u$, we first compute a embedding $\mathbf{v}_t$ for visit~$t$:
\begin{align}
\label{eq:visit_emb}
    \mathbf{v}_t = \frac{1}{\mid C_t \mid}\sum_{c_i \in C_t}{\mathbf{H}_c^i} \in \mathbb{R}^{{d^{(L)}_c}}.
\end{align}
After the collaborative graph learning, $\mathbf{H}_c^i$ contains the information of its multi-hop neighbor diseases by the connection of patient nodes. Hence, different from {\gram}, it enables the model to effectively predict diseases that have never been diagnosed on a patient before. We then employ GRU on $\mathbf{v}_t$ to learn visit temporal features and get a hidden representation $\mathbf{R} = \{\mathbf{r}_1, \mathbf{r}_2, \dots, \mathbf{r}_T\}$ where the size of the RNN cell is $h$:
\begin{align}
    \label{eq:gru1}
    \mathbf{R} = \mathbf{r}_1, \mathbf{r}_2, \dots, \mathbf{r}_T &= \text{GRU}(\mathbf{v}_1, \mathbf{v}_2, \dots, \mathbf{v}_T) \in \mathbb{R}^{T \times h},
\end{align}
% It can also be replaced by other RNN modules such as a basic RNN cell and LSTM.
Then we apply a location-based attention~\cite{luong2015effective} to calculate the final hidden representation $\mathbf{o}_v$ of all visits:
\begin{align}
    \label{eq:attention1}
    \boldsymbol{\alpha} &= \text{softmax}\left( \mathbf{R}\mathbf{w}_{\alpha} \right) \in \mathbb{R}^{T}, \\
    \mathbf{o}_v &= \boldsymbol{\alpha}\mathbf{R} \in \mathbb{R}^h, \label{eq:attention3}
\end{align}
where $\mathbf{w}_{\alpha} \in \mathbb{R}^{h}$ is a context vector for attention and $\boldsymbol{\alpha}$ is the attention weight for each visit.

\subsubsection{Guiding Attention on Clinical Notes}
We incorporate the clinical notes $N_T$ from the latest visit $V_T$, since $N_T$ generally contains the medical history and future plan for a patient. We  propose an attention regulation strategy that automatically highlights key words, considering traditional attention mechanisms in NLP have raised concerns as a tool to audit a model~\cite{jain-wallace-2019-attention,serrano-smith-2019-attention}. Pruthi \textit{et al.}~\cite{pruthi-etal-2020-learning} present a manipulating strategy using a set of pre-defined impermissible tokens and penalizing the attention weights on these impermissible tokens. To implement the regulation strategy, we propose a TF-IDF-rectified attention method on clinical notes. Regarding all patients' notes as a corpus and each patient's note as a document, for a patient $u$, we first calculate the TF-IDF weight $\beta_{i}$ for each word $\omega_i$ in $u$'s note $N_T$ and normalize the weights into [0, 1]. Then, we select the embedding $\mathbf{q}_i \in \mathbb{R}^{d_w}$ from a randomly initialized word embedding matrix $\mathbf{Q} \in \mathbb{R}^{|\mathcal{N}| \times d_w}$. For attention in Eq.~(\ref{eq:attention1}), the context vector $\mathbf{w}_{\alpha}$ is randomly initialized, while clinical notes are correlated  with diagnoses. Therefore, we adopt $\mathbf{o}_v$ as the context vector. Firstly, we project word embeddings $\mathbf{Q}$ into the dimension of visits to multiply the context vector $\mathbf{o}_v$:
\begin{align}
    \mathbf{Q}' = \mathbf{Q}\mathbf{W}_q \in \mathbb{R}^{|\mathcal{N}| \times h}
\end{align}
Then, let $\mathbf{N}$ be the embedding matrix selected from $\mathbf{Q}'$ for words in $N_T$, we calculate the attention weight $\boldsymbol{\alpha}'$ as well as the output ${\mathbf{o}_n}$ for clinical~notes:
\begin{align}
    \boldsymbol{\alpha}' &= \text{softmax}\left( \mathbf{N}\mathbf{o}_v \right) \in \mathbb{R}^{|N_T|}, \\
    \mathbf{o}_n &= \boldsymbol{\alpha}'\mathbf{N} \in \mathbb{R}^{h}.
\end{align}
For a word with a high TF-IDF weight in a clinical note, we expect the model to focus on this word with a high attention weight. Therefore, we introduce a TF-IDF-rectified attention penalty $\mathcal{L}_0$ for the attention weights of words:
\begin{align}
    \mathcal{L}_0 = -\sum_{\omega_i \in N_T}{\left(\alpha'_i \log\beta_{i} + (1 - \alpha'_i)\log{(1 - \beta_{i})}\right)}.
\end{align}
The attention weights that mismatch the TF-IDF weights will be penalized. We believe that irrelevant (impermissible) words such as ``patient'' and ``doctor'' tend to have low TF-IDF weights. Finally, we concatenate ${\mathbf{o}_n}$ and $\mathbf{o}_v$ as the output ${\mathbf{O}} \in \mathbb{R}^{2h}$ for patient $u$: ${\mathbf{O}} = \mathbf{o}_v \oplus {\mathbf{o}}_n$.

\subsubsection{Prediction and Inference}
Diagnosis prediction is a multi-label classification task, while heart failure prediction is a binary classification task. We both use a dense layer with a sigmoid activation function on the model output $\mathbf{O}$ to calculate the predicted probability $\hat{\mathbf{y}}$. The loss function of classification for both tasks is cross-entropy loss $\mathcal{L}_c$. Then, we combine the TF-IDF-rectified penalty $\mathcal{L}_0$ and cross-entropy loss as the final loss $\mathcal{L}$ to train the model:
\begin{align}
    \label{eq:model_loss}
    \mathcal{L} = \lambda\mathcal{L}_0 + \text{CrossEntropy}(\hat{\mathbf{y}}, \mathbf{y}).
\end{align}
Here, $\mathbf{y}$ is the ground-truth label of medical codes or heart failure, and $\lambda$ is a coefficient to adjust $\mathcal{L}_0$. In the inference phase, we freeze the trained model and retrieve the embeddings $\mathbf{H}_c$ of medical codes at the output of heterogeneous graph learning. Then, given a new patient for inference, we continue from Eq.~(\ref{eq:visit_emb}) and make predictions. 

\begin{table}
\small
    \centering
    \begin{tabular}{lr}
        \toprule
        Patient number & 7,125\\
        Avg. visit number per patient & 2.66\\
        Patient number with heart failure & 2,604 \\
        \midrule
        Medical code (disease) number & 4,795\\
        Avg. code number per visit & 13.27\\
        \midrule
        Dictionary size in notes & 67,913 \\
        Avg. word number per note & 4,732.28 \\
        \bottomrule
    \end{tabular}
    \caption{Statistics of the MIMIC-III dataset.}
    \label{tab:dataset}
\end{table}
\section{Experiments}
\subsection{Experimental Setup}
\label{sec:experiments}
\subsubsection{Dataset Description}

We use the MIMIC-III dataset~\cite{johnson2016mimic} to evaluate \modelname. Table~\ref{tab:dataset} shows the basic statistics of MIMIC-III. We select patients with multiple visits (\# of visits $\ge$ 2) and select clinical notes except the type of ``Discharge summary'', since it has a strong indication to predictions and is unfair to be used as features. For each note, we use the first 50,000 words, while the rest are cut off for computational efficiency, given the average word number per note is less than 5,000. We split MIMIC-III randomly according to patients into training/validation/test sets with patient numbers as 6000/125/1000. We use the codes in patients' last visit as labels and other visits as features. For heart failure prediction, we set labels as 1 if patients are diagnosed with heart failure in the last visit. Finally, the observation graph is built based on the training set. A 5-level hierarchical structure and the ontology graph are built according to ICD-9-CM.

\subsubsection{Evaluation Metrics}
We adopt weighted $F_1$ score (w-$F_1$~\cite{bai2018interpretable}) and top $k$ recall (R@$k$~\cite{choi2016doctor}) for diagnosis predictions. w-$F_1$ is a weighted sum of $F_1$ for each class. R@$k$ is the ratio of true positive numbers in top $k$ predictions by the total number of positive samples, which measures the prediction performance on a subset of classes. For heart failure predictions, we use $F_1$ and the area under the ROC curve (AUC), since it is a binary classification on imbalanced test data.

\subsubsection{Baselines}
To compare \modelname~with state-of-the-art models, we select the following models as baselines: 1) \textit{RNN-based models}: \retain~\cite{choi2016retain}, \dipole~\cite{ma2017dipole}, \timeline~\cite{bai2018interpretable}; 2) \textit{CNN-based models}: \deepr~\cite{Phuoc2017deepr}; 3) Graph-based models\textit{}: \gram~\cite{choi2017gram}, \medgcn~\cite{mao2019medgcn}; and 4) \textit{A logistic regression model}, {\notes}, on clinical notes using only TF-IDF features of each note (whose dimension is the dictionary size).

Deepr, GRAM, and Timeline use medical code embeddings as inputs, while others use multi-hot vectors of medical codes. We do not consider SMR~\cite{wang2017safe}  because 1) it does not compare with the above state-of-the-art models and 2) it focuses on medication recommendation which is different from our tasks. We also do not compare with MiME~\cite{choi2018mime} and GCT~\cite{gct_aaai20} because we do not use treatments and lab results in our data.

\subsubsection{Parameters}
We randomly initialize embeddings for diseases, patients, and clinical notes and select the paramters by a grid search. The embedding sizes $d_c$, $d_p$, and $d_w$ are 32, 16, and 16. The graph layer number $L$ is 2. The hidden dimensions {\small$d_p^{(1)}$, $d_c^{(1)}$, and $d_c^{(2)}$} are 32, 64, and 128, and the GRU unit $h$ is set to 200. The coefficient $\lambda$ in $\mathcal{L}_0$ for diagnosis and heart failure prediction is 0.3 and 0.1. We set the learning rate as $10^{-3}$, optimizer as Adam, and use 200 epochs for training. The source code of {\modelname} is released at \href{https://github.com/LuChang-CS/CGL}{https://github.com/LuChang-CS/CGL}.
\begin{table}
    \centering
   \scalebox{0.85}{\begin{tabular}{lcccc}
        \toprule
         \multirow{1}{*}{\textbf{Models}} & \multirow{1}{*}{\textbf{w-}$\boldsymbol{F_1}$  (\%)} &  \textbf{R@20}  (\%) &  \textbf{R@40}  (\%) &
        \multirow{1}{*}{ \textbf{Param.}} \\
        \midrule
        \retain & 19.66 (0.58) & 33.90 (0.47) & 42.93 (0.39) & 2.90M \\
        \deepr & 12.38 (0.01) & 28.15 (0.08) & 37.26 (0.14) & 0.80M \\
        \gram & 21.06 (0.19) & 36.37 (0.16) & 45.61 (0.27) & 1.38M \\
        \dipole & 11.24 (0.19) & 26.96 (0.15) & 36.83 (0.26) & 2.08M \\
        \timeline & 16.83 (0.62) & 32.08 (0.66) & 41.97 (0.74) & 1.23M \\
        \medgcn & 20.93 (0.25) & 35.69 (0.50) & 43.36 (0.46) & 4.59M \\
        LR$_{\text{notes}}$ & 17.56 (0.41) & 36.71 (0.28) & 46.02 (0.38) & 325.65M \\
        \midrule
         \modelname & \textbf{22.97 (0.19)} & \textbf{38.19 (0.16)} & \textbf{48.26 (0.15)} & 3.55M \\
         \bottomrule
    \end{tabular}}
    \caption{Diagnosis prediction results in w-${F_1}$ and R@${k}$.}
    \label{tab:result_code}
\end{table}

\subsection{Experimental Results}
\subsubsection{Diagnosis and Heart Failure Prediction}

Table~\ref{tab:result_code} shows the results of baselines and \modelname~on diagnosis prediction. We use $ k = [20, 40]$ for R@$k$. Each model is trained for 5 times with different variable initializations. The mean and standard deviation are reported. The proposed \modelname~model outperforms all the baselines. We think this is mostly because {\modelname~} captures hidden connections of patients and diseases and utilizes clinical notes. In addition, the results of \notes~indicate that only using clinical notes does not improve performance in predicting diagnosis. Table~\ref{tab:result_hf} shows the heart failure prediction results. We observe that \modelname~also achieves the best performance in terms of AUC and $F_1$. 

\begin{table}
    \centering
    \scalebox{0.85}{\begin{tabular}{lccc}
        \toprule
        \textbf{Models} & \textbf{AUC} (\%) & $\boldsymbol{F_1}$ (\%) & \textbf{Param.} \\
        \midrule
        \retain & 82.73 (0.21) & 71.12 (0.37) & 1.67M \\
        \deepr & 81.29 (0.01) & 68.42 (0.01) & 0.49M \\
        \gram & 82.82 (0.06) & 71.43 (0.05) & 0.76M \\
        \dipole & 81.66 (0.07) & 70.01 (0.04) & 1.45M \\
        \timeline & 80.75 (0.46) & 69.81 (0.34) & 0.95M \\
        \medgcn & 81.25 (0.15) & 70.86 (0.18) & 3.98M \\
        LR$_{\text{notes}}$ & 80.33 (0.12) & 69.18 (0.27) & 0.07M \\
        \midrule
        \modelname & \textbf{85.66 (0.19)} & \textbf{72.68 (0.22)} & 1.62M \\
        \bottomrule
    \end{tabular}}
    \caption{Heart failure prediction results in AUC and $F_1$.}
    \label{tab:result_hf}
\end{table}

\begin{table}
    \centering
    \scalebox{0.85}{\begin{tabular}{lcccccc}
        \toprule
        \multirow{2}{*}{\textbf{Models}} & \multicolumn{3}{c}{\textbf{Diagnosis}} &\multicolumn{3}{c}{\textbf{Heart failure}}\\
        \cmidrule{2-4}  \cmidrule{5-7}
        & w-$F_1$ & R@20 & Param. & AUC & $F_1$ & Param. \\
        \midrule
        \modelname $_{h\text{-}}$ & 20.87 & 35.66 & 3.98M & 82.58 & 71.02 & 2.04M \\
        \modelname $_{n\text{-}}$ & 22.10 & 37.59 & 1.50M & 84.53 & 71.96 & 0.53M \\
        \modelname $_{w\text{-}}$ & 22.06 & 37.31 & 3.54M & 83.91 & 71.59 & 1.60M \\
        \modelname & \textbf{22.97} & \textbf{38.19} & 3.55M & \textbf{85.66} & \textbf{72.68} & 1.62M \\
        \bottomrule
    \end{tabular}}
    \caption{w-$F_1$, R@20 of diagnosis prediction and AUC, $F_1$ of heart failure prediction for {\modelname} variants. \modelname$_{h\text{-}}$: no hierarchical embedding; \modelname$_{n\text{-}}$: no clinical notes; \modelname$_{w\text{-}}$: no ontology weights.}
    \label{tab:ablation}
\end{table}

\subsubsection{Ablation Study}
To study the effectiveness of components, we also compare 3 {\modelname} variants: \modelname~without hierarchical embedding (\modelname$_{h\text{-}}$), \modelname~without clinical notes as inputs (\modelname$_{n\text{-}}$), and \modelname~without ontology weights (\modelname$_{w\text{-}}$). The results are shown in Table~\ref{tab:ablation}. We observe that even without clinical notes, {\modelname $_{n\text{-}}$} with hierarchical embeddings and ontology weights still achieves the best performance among all other baselines. This indicates that domain knowledge including hierarchical embeddings and ontology weights also help to learn better representations of medical codes. In addition, from Table~\ref{tab:ablation} we can infer that the complexity of {\modelname} mostly comes from modeling clinical notes, i.e., word embeddings. Therefore, {\modelname} is scalable and can be generalized to other tasks when clinical notes are not accessible.

\subsubsection{Prediction Analysis}
\paragraph{New-onset diseases.}
For a patient, new-onset diseases denote new diseases in future visits which have not occurred in previous visits of this patient. We use the ability of predicting new-onset diseases to measure learned diagnostic similarity of patients. It is natural for a model to predict diseases that have occurred in previous visits. With the help of other similar patients' records, the model should be able to predict new diseases for a patient. The idea is similar to collaborative filtering in recommender systems. If two patients are similar, one of them may be diagnosed with new-onset diseases which have occurred in the other patient. We also use R@$k$ ($k=[20, 40]$) to evaluate the ability of predicting occurred and new-onset diseases. Here, R@$k$ denotes the ratio between the number of correctly predicted occurred (or new) diseases and the number of ground-truth diseases. We select GRAM and MedGCN which have good performance in diagnosis prediction, and {\modelname $_{n\text{-}}$} without clinical notes, because we want to explore the effectiveness of the proposed observation and ontology graphs. Table~\ref{tab:occurred} shows the results of R@$k$ on test data. We can see that {\modelname $_{n\text{-}}$} has similar results to {\gram} on occurred diseases while achieving superior performance on new-onset diseases. This verifies that our proposed collaborative graph learning is able to learn from similar patients and predict new-onset diseases in the future.

\begin{table}
    \centering
    \scalebox{0.85}{\begin{tabular}{lcccc}
        \toprule
        \multirow{2}{*}{\textbf{Models}} & \multicolumn{2}{c}{\textbf{Occurred}} &\multicolumn{2}{c}{\textbf{New-onset}}\\
        \cmidrule{2-3}  \cmidrule{4-5}
        & R@20 & R@40 & R@20 & R@40 \\
        \midrule
        \gram   & 21.05 & 23.11 & 15.32 & 22.50 \\
        \medgcn & 20.51 & 21.89 & 15.38 & 21.53 \\
        \modelname $_{n\text{-}}$ & \textbf{21.26} & \textbf{23.85} & \textbf{16.33} & \textbf{23.58} \\
        \bottomrule
    \end{tabular}}
    \caption{R@$k$ of predicting occurred/new-onset diseases.}
    \label{tab:occurred}
\end{table}

\begin{figure}
    \centering
    \scalebox{0.85}{\begin{subfigure}[t]{0.32\linewidth}
        \includegraphics[width=\linewidth]{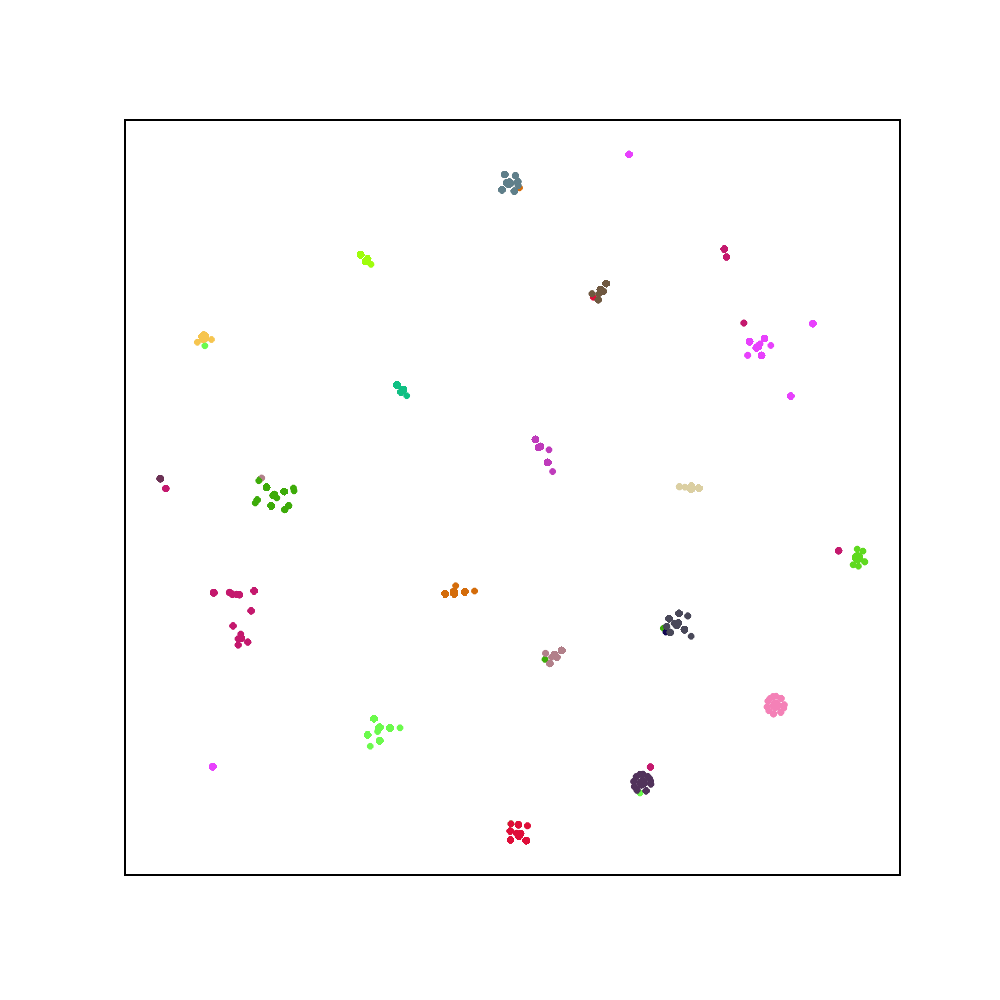}
        \subcaption{GRAM level 1}
    \end{subfigure}}
    \scalebox{0.85}{\begin{subfigure}[t]{0.32\linewidth}
        \includegraphics[width=\linewidth]{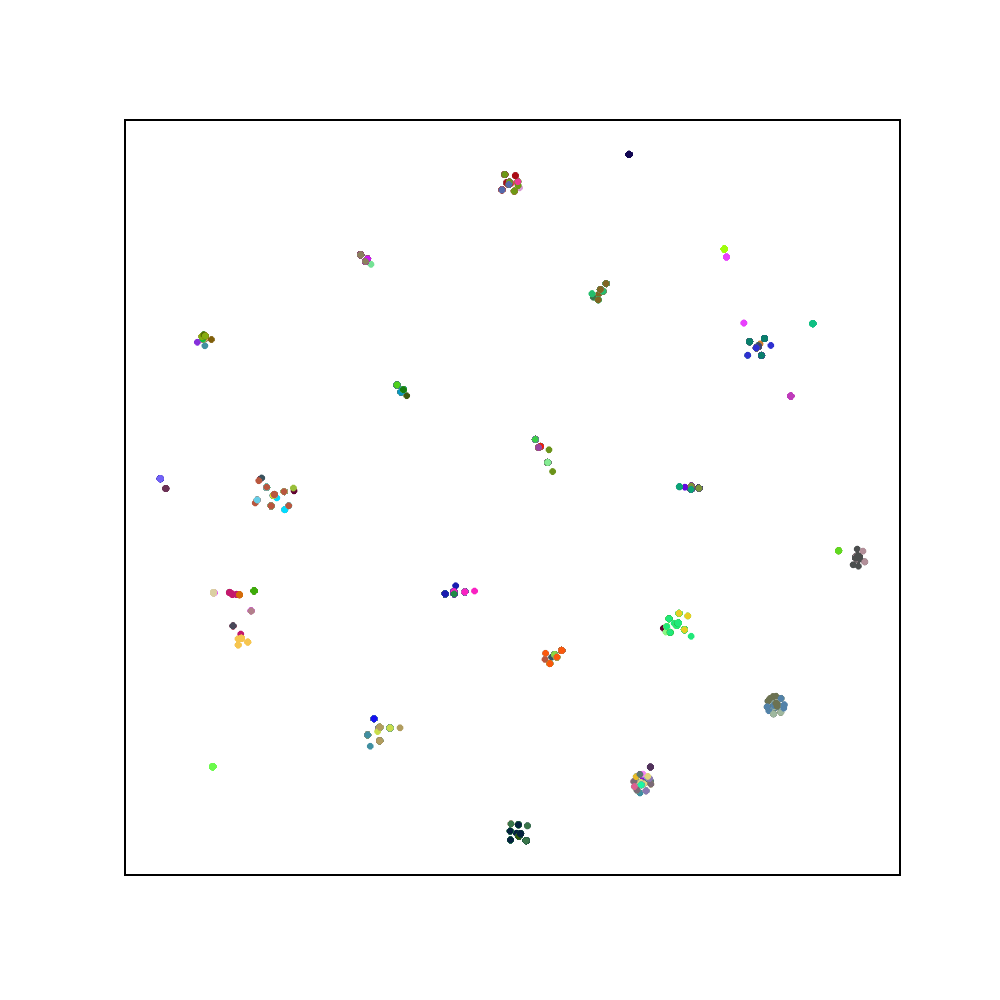}
        \subcaption{GRAM level 2}
    \end{subfigure}}
    \scalebox{0.85}{\begin{subfigure}[t]{0.32\linewidth}
        \includegraphics[width=\linewidth]{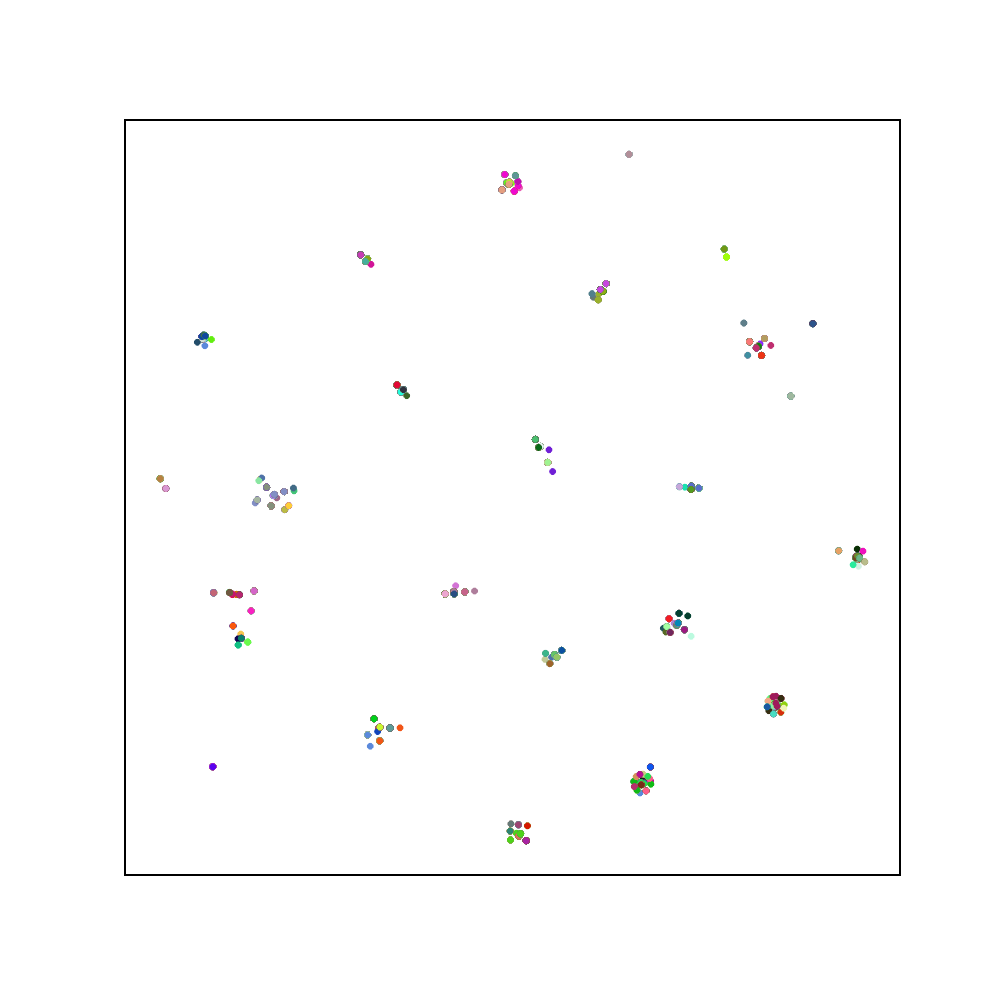}
        \subcaption{GRAM level 3}
    \end{subfigure}}
    \par\bigskip
    \scalebox{0.85}{\begin{subfigure}[t]{0.32\linewidth}
        \includegraphics[width=\linewidth]{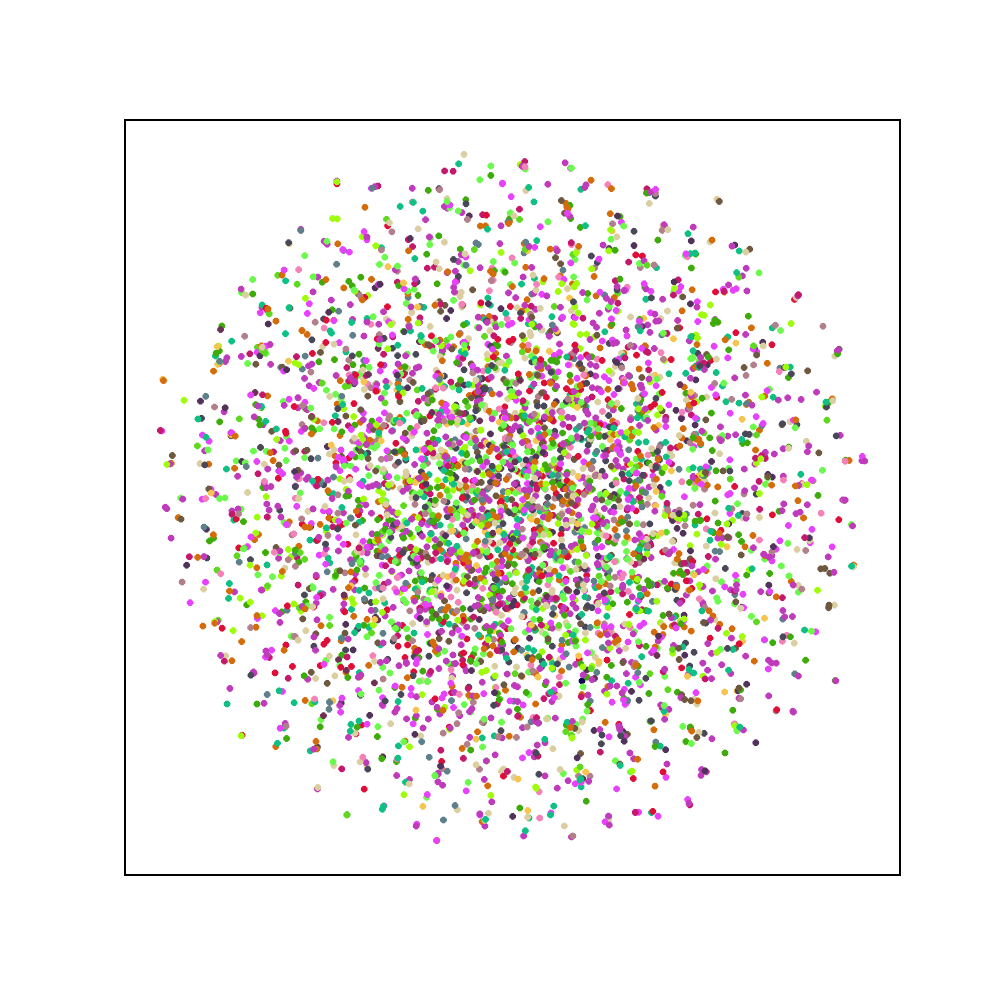}
        \subcaption{Timeline level 1}
    \end{subfigure}}
    \scalebox{0.85}{\begin{subfigure}[t]{0.32\linewidth}
        \includegraphics[width=\linewidth]{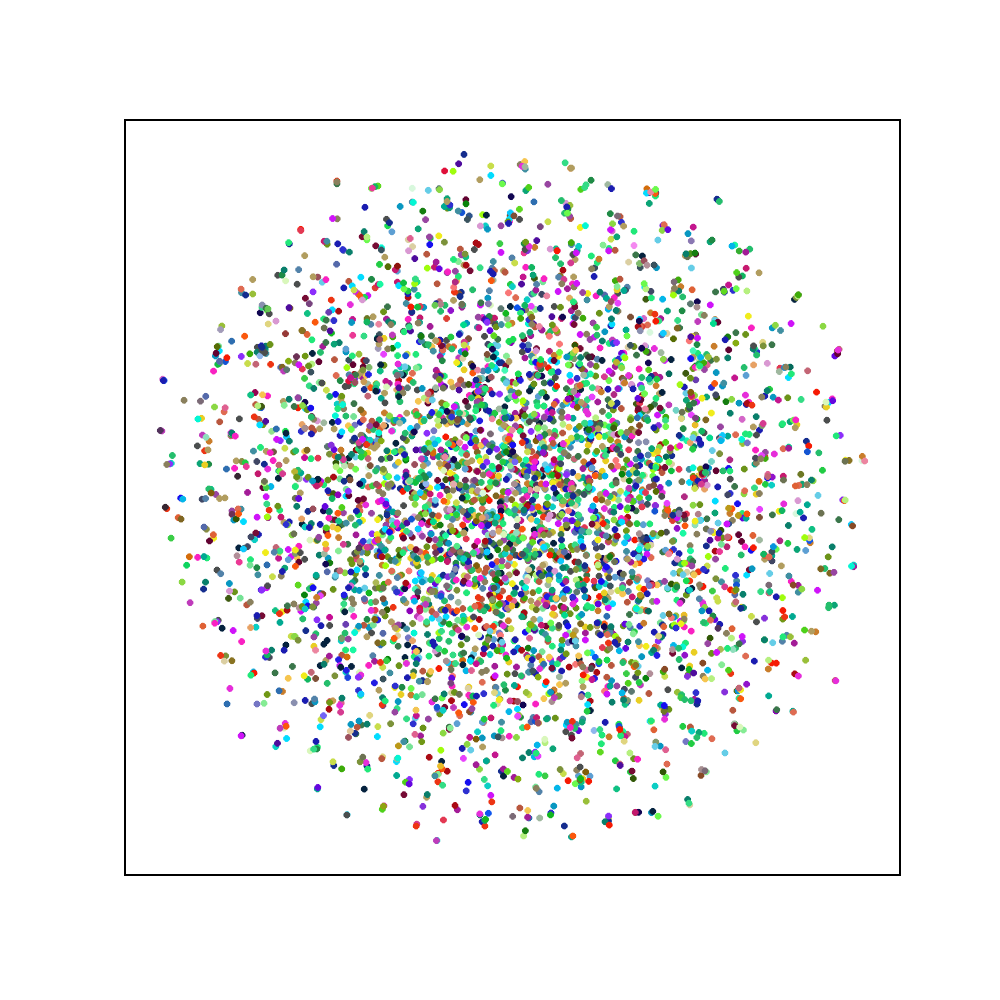}
        \subcaption{Timeline level 2}
    \end{subfigure}}
    \scalebox{0.85}{\begin{subfigure}[t]{0.32\linewidth}
        \includegraphics[width=\linewidth]{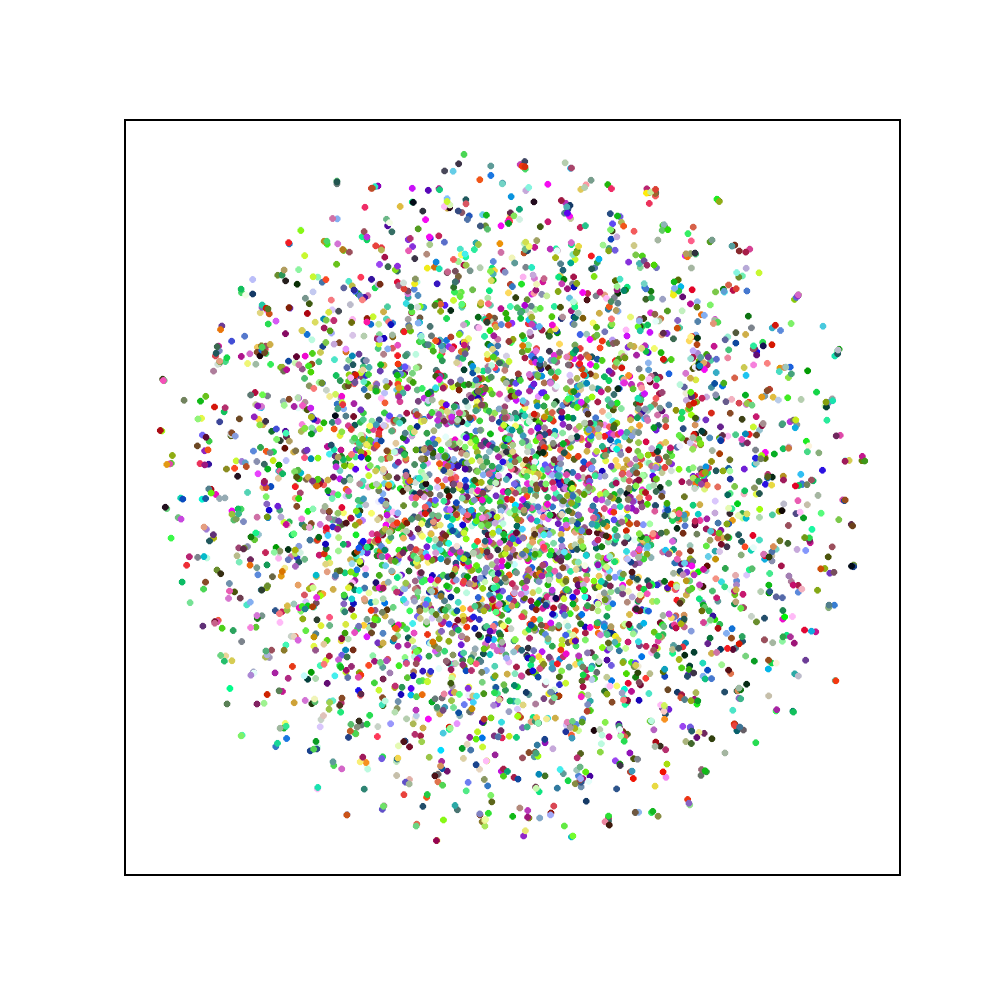}
        \subcaption{Timeline level 3}
    \end{subfigure}}
    \par\bigskip
    \scalebox{0.85}{\begin{subfigure}[t]{0.32\linewidth}
        \includegraphics[width=\linewidth]{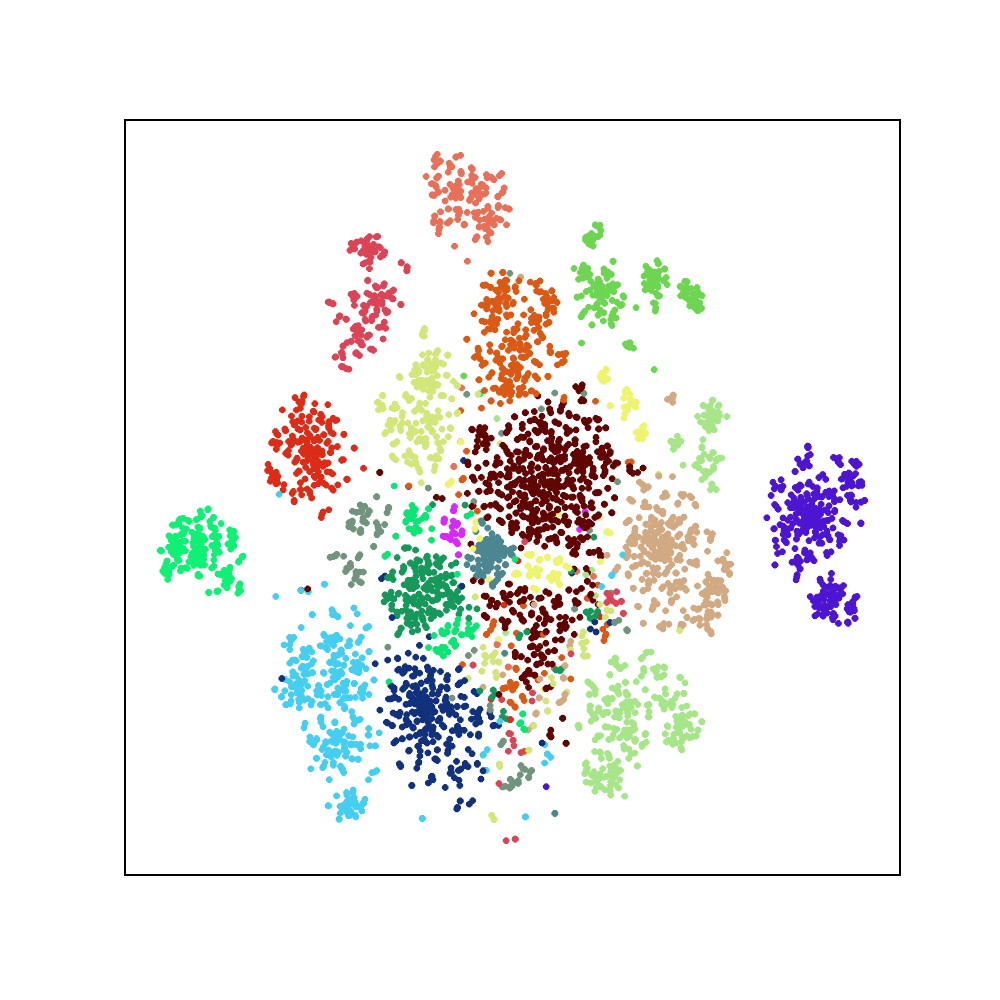}
        \subcaption{\modelname~level 1}
    \end{subfigure}}
    \scalebox{0.85}{\begin{subfigure}[t]{0.32\linewidth}
        \includegraphics[width=\linewidth]{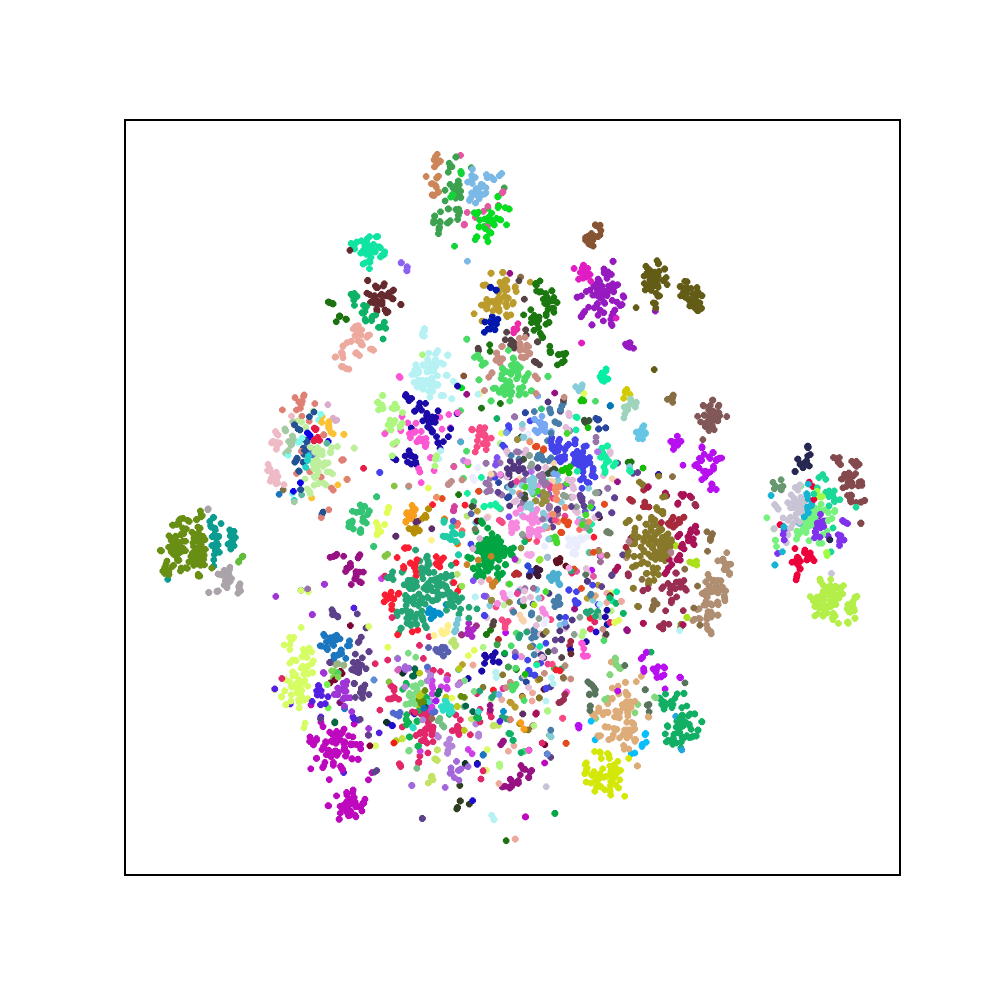}
        \subcaption{\modelname~level 2}
    \end{subfigure}}
    \scalebox{0.85}{\begin{subfigure}[t]{0.32\linewidth}
        \includegraphics[width=\linewidth]{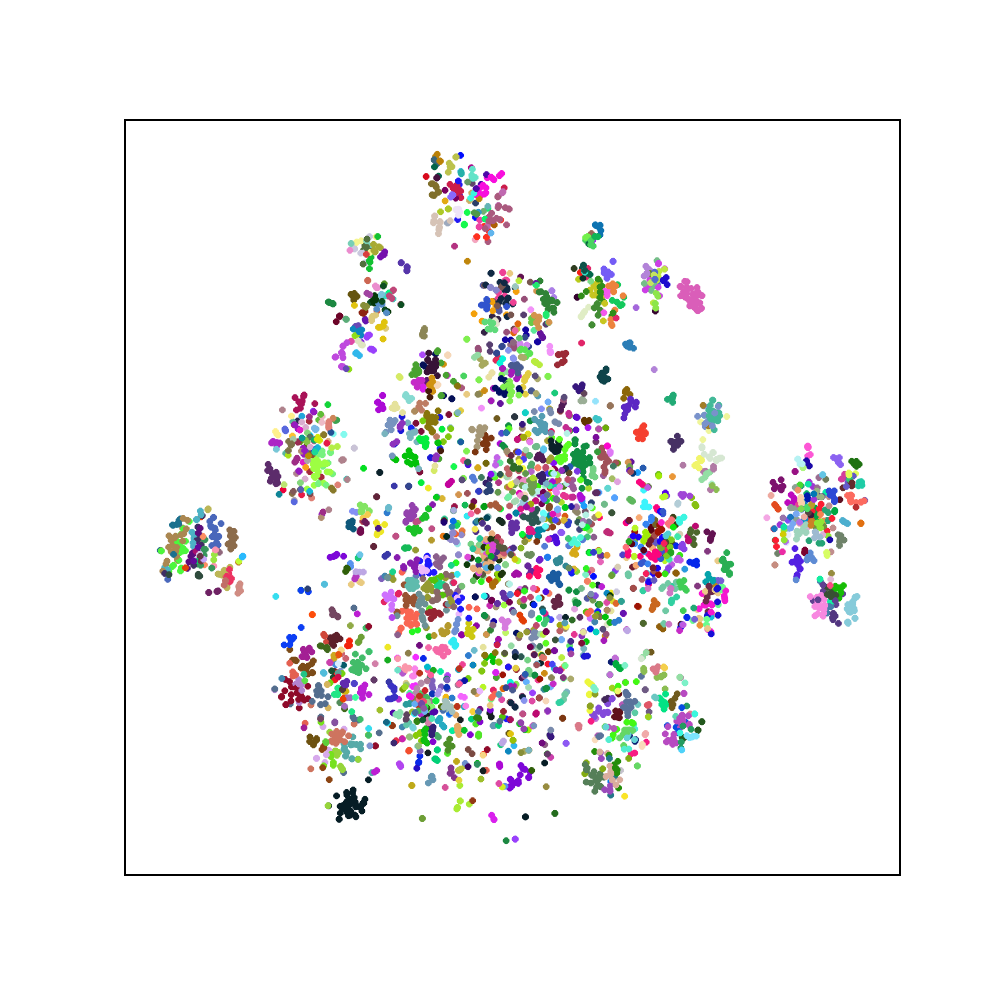}
        \subcaption{\modelname~level 3}
    \end{subfigure}}
  \caption{Code embeddings in 3 levels learned by GRAM, Timeline, and \modelname. Colors correspond to disease types in each level.}
  \label{fig:code_emb}
\end{figure}

\paragraph{Disease embeddings.}
To show the similarity of diseases, we plot the learned 4795 code embeddings $\mathbf{H}_c$ using t-SNE \cite{maaten2008visualizing}. \figurename{~\ref{fig:code_emb}} shows the embeddings learned by GRAM, Timeline, and \modelname~in 3 levels. Colors denotes different disease types in each level. In \figurename{~\ref{fig:code_emb}}, disease embeddings learned by GRAM and {\modelname} are basically clustered according to their real categories, while Timeline seems like a random distribution. In the plot of GRAM, we observe the clusters are far away from each other given large inter-cluster distances, while nodes in a cluster are close to each other due to small intra-cluster distances. We can observe that the embeddings learned by GRAM do not capture distinguishable features of low-level diseases as well as the relationships across clusters. Therefore, we can infer that learning proper representations that reflect disease hierarchical structures and correlations is helpful for predictions.

\paragraph{Contribution of notes.}
We compare the proposed TF-IDF rectified attention weights with regular attention weights to verify if the model focuses on important words. Table~\ref{tab:contrib_note} demonstrates an example with a part of a note and predicted diagnoses. In this example, the patient is diagnosed with 33 diseases, and \modelname~predicts 10 of them correctly in top 20 predicted codes. Important words with high $\alpha'$ values are highlighted in pink. We first observe that pink words are relevant to diagnoses. In addition, we notice the rectified attention weights are more semantically interpretable. For example, ``acute'' and ``HCAP'' (Health care-associated pneumonia) get higher weights with the rectified attention loss. Meanwhile, we show the unimportant words with low $\alpha'$ values in gray. We observe that our model detects unimportant words which have less contributions. For example, ``patient'' and ``diagnosis'' are regarded as an unimportant word but not captured in the regular attention mechanism. Therefore, we may conclude that the TF-IDF-rectified attention method improves the accuracy of interpretations using clinical notes.

\newcommand{\cbx}[2]{\setlength{\fboxsep}{0.5pt}\colorbox{#1}{#2}}

\begin{table}
\small
    \centering
    \scalebox{0.75}{\begin{tabular}{p{0.41\linewidth}|p{0.41\linewidth}|p{0.36\linewidth}}
        \toprule
        \textbf{Without penalty} & \textbf{With penalty} &\textbf{ Correct Predictions} \\\midrule 
        ... Patient had fairly acute \cbx{pink}{decompensation} of \cbx{lightgray}{respiratory} status today with \cbx{pink}{hypoxia} and \cbx{pink}{hypercarbia} associated with \cbx{pink}{hypertension} ... Differential diagnosis includes \cbx{lightgray}{flash} \cbx{pink}{pulmonary} \cbx{pink}{edema}~and acute exacerbation of CHF vs aspiration vs infection (HCAP) ... Acuity suggests possible \cbx{lightgray}{flash} \cbx{pink}{pulmonary} \cbx{pink}{edema} vs aspiration ...
        & ... \cbx{lightgray}{Patient} had fairly \cbx{pink}{acute} \cbx{pink}{decompensation} of \cbx{pink}{respiratory} status today with \cbx{pink}{hypoxia} and \cbx{pink}{hypercarbia} associated with \cbx{pink}{hypertension} ... Differential \cbx{lightgray}{diagnosis} includes flash \cbx{pink}{pulmonary} \cbx{pink}{edema} and \cbx{pink}{acute} exacerbation of CHF vs aspiration vs infection (\cbx{pink}{HCAP}) ... Acuity suggests possible flash \cbx{pink}{pulmonary} \cbx{pink}{edema} vs aspiration ...
        & {\begin{itemize}[leftmargin=*]
            \item {Hypertensive chronic kidney disease}
            \item {Acute respiratory failure}
            \item {Congestive heart failure}
            \item {Diabetes}
            \item ...
        \end{itemize}} \\
        \bottomrule
    \end{tabular}}
    \caption{An example of word contributions without/with the TF-IDF rectified penalty. The pink/gray color denotes high/low attention weights. }
    \label{tab:contrib_note}
\end{table}

\section{Conclusion}
\label{sec:conclusion}
In this paper, we propose \modelname, a collaborative graph learning model to jointly learn the representations of patients and diseases, and effectively utilize clinical notes in EHR data. We conducted experiments on real-world EHR data to demonstrate the effectiveness of the learned representations and performance improvements of {\modelname} over state-of-the-art models. We also provide analysis of {\modelname} on multiple aspects, including new onset diseases, disease embeddings, and contribution of clinical notes. In the future, we plan to explore methods to quantify the contributions of certain admissions to each predicted medical code. 
Usage of single admission records in EHR data will also be considered for further investigation.

\section*{Acknowledgments}
This work was supported in part by US National Science
Foundation under grants 1838730 and 1948432.
SK was supported in part by the NLM of the NIH under Award Number R01LM013308.
\bibliographystyle{named}
\bibliography{ref}

\end{document}